\pdfoutput=1

\documentclass[11pt]{article}
\usepackage{amsmath}
\usepackage{booktabs}
\usepackage[preprint]{acl}

\usepackage{times}
\usepackage{latexsym}
\usepackage{amssymb}
\usepackage{adjustbox}
\usepackage{multirow}
\usepackage{algorithm}
\usepackage{algpseudocode}
\usepackage{caption}
\usepackage{subcaption}
\usepackage{listings}
\usepackage{hyperref}
\usepackage{adjustbox}
\usepackage{xcolor}

\newcommand{\by}{\mathbf{y}}
\newcommand{\bx}{\mathbf{x}}

\usepackage[T1]{fontenc}

\usepackage[utf8]{inputenc}

\usepackage{microtype}

\usepackage{inconsolata}

\usepackage{graphicx}

\title{Effective Reinforcement Learning for Reasoning in Language Models}

\author{Lianghuan Huang\thanks{The first two authors are listed alphabetically. Correspondence to: Osbert Bastani <obastani@seas.upenn.edu>.}\thanks{University of Pennsylvania, Philadelphia (PA),
US} \And Shuo Li$^{*\dag}$ \And Sagnik Anupam$^\dag$ \AND Insup Lee$^\dag$ \And Osbert Bastani$^{\dag}$
  }

\begin{document}
\maketitle
\begin{abstract}
Reinforcement learning (RL) has emerged as a promising strategy for improving the reasoning capabilities of language models (LMs) in domains such as mathematics and coding. However, most modern RL algorithms were designed to target robotics applications, which differ significantly from LM reasoning. We analyze RL algorithm design decisions for LM reasoning, for both accuracy and computational efficiency, focusing on relatively small models due to computational constraints. Our findings are: (i) on-policy RL significantly outperforms supervised fine-tuning (SFT), (ii) PPO-based off-policy updates increase accuracy instead of reduce variance, and (iii) removing KL divergence can lead to more concise generations and higher accuracy. Furthermore, we find that a key bottleneck to computational efficiency is that the optimal batch sizes for inference and backpropagation are different. We propose a novel algorithm, DASH, that performs \textit{preemptive sampling} (i.e., sample a large batch and accumulate gradient updates in small increments), and \textit{gradient filtering} (i.e., drop samples with small advantage estimates). We show that DASH reduces training time by 83\% compared to a standard implementation of GRPO without sacrificing accuracy. Our findings provide valuable insights on designing effective RL algorithms for LM reasoning.\footnote{Our code is here:  \url{https://github.com/shuoli90/efficient_reasoning}.}
\end{abstract}

\section{Introduction}

Recent advancements have shown that reinforcement learning (RL) algorithms can significantly enhance the mathematical reasoning capabilities of language models (LMs)~\citep{deepseekai2025deepseekr1incentivizingreasoningcapability, qwen2025qwen25technicalreport, zeng2025simplerlzooinvestigatingtamingzero}. 
Despite these results, there has been little systematic understanding of how different RL design decisions contribute to their effectiveness in the LM reasoning setting. Many of these algorithms were originally designed for robotics, while LM reasoning exhibits qualitatively different learning patterns, meaning different design decisions may be more effective~\citep{ahmadian2024basicsrevisitingreinforcestyle}; indeed, even the space of relevant design decisions may be different for LM reasoning compared to robotics. Our goal is to answer the following question:

\textbf{How do we design effective RL algorithms for improving the reasoning capabilities of LMs?}

\begin{figure}[t]
\centering
\includegraphics[width=0.5\textwidth]{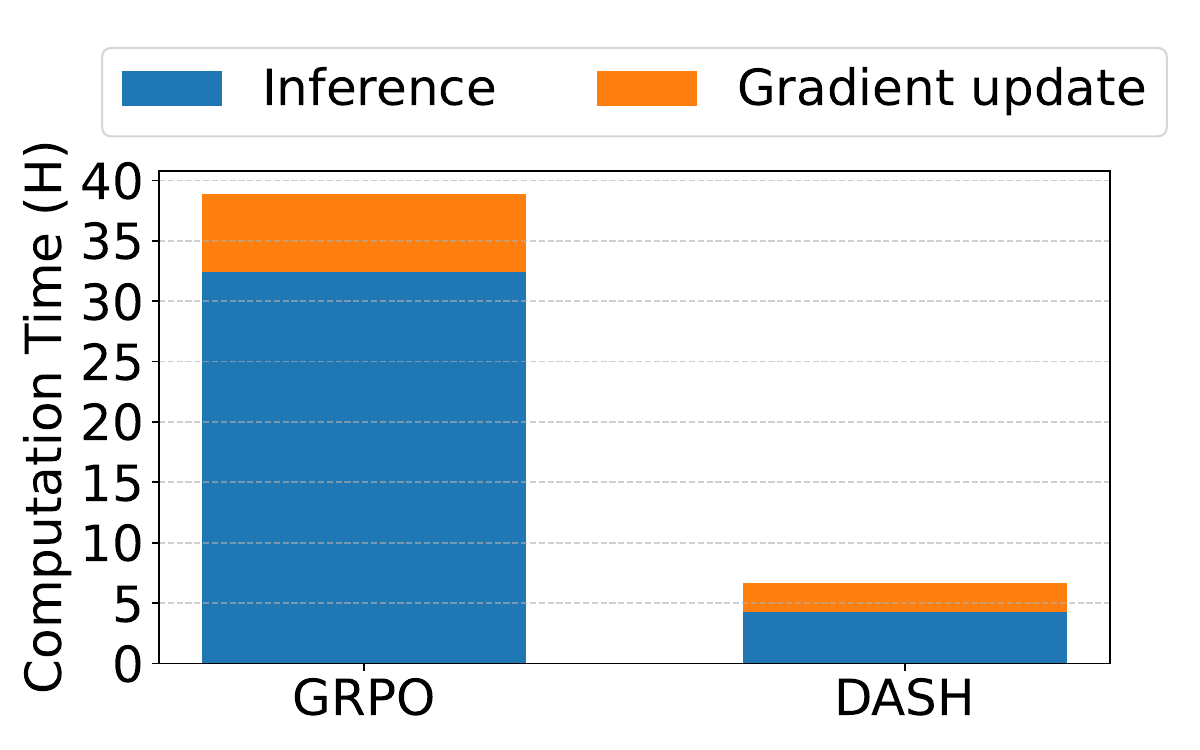}
\caption{DASH can reduce running time by 83\% compared to GRPO by using \textit{preemptive sampling} (Section~\ref{sec:preemptive_sampling}) and \emph{gradient filtering} (Section~\ref{sec:gf}).}
\label{fig:time comparison}
\end{figure}

Importantly, we are interested not only in the performance (i.e., the final accuracy), but also efficiency (i.e., how quickly the algorithm converges). Furthermore, we focus on relatively small models (0.5B, 1.5B, and 3B) where we can explore a variety of different RL algorithms.

We perform a systematic analysis of the different design decisions in an RL algorithm. We start by considering the two most prevalent types of algorithms: supervised fine-tuning (SFT)~\citep{chen2023fireactlanguageagentfinetuning, zeng2023agenttuningenablinggeneralizedagent}, also known as behavior cloning, and on-policy RL (e.g., policy gradient~\citep{NIPS1999_464d828b}, PPO~\citep{schulman2017trustregionpolicyoptimization}, GRPO~\citep{shao2024deepseekmathpushinglimitsmathematical}, etc.). While SFT is much more efficient, we find it to be significantly less effective at improving reasoning ability for the models we consider; this may be due to the inability for smaller models to effectively mimic the reasoning traces of larger models or humans. In contrast, we find that on-policy RL is highly effective at improving performance.

Next, we compare different kinds of on-policy RL algorithms. Compared to the original policy gradient (PG) algorithm, PPO is designed to improve stability by ``freezing'' the inference policy and taking multiple gradient steps. We find that while PPO can improve accuracy, it has significantly higher variance compared to PG, which is the opposite of conventional wisdom. PPO also introduces a KL divergence term to regularize the training policy towards the inference policy; however, we find that it leads to lengthier generations and worse performance.

While on-policy RL is highly effective, existing algorithms are computationally expensive to run. Analyzing their performance bottlenecks, we find that the sampling procedure is a key bottleneck. The key issue is that inference and training require significantly different batch sizes to make optimal use of computational resources. Thus, it is much more effective to perform inference in a single large batch, and then accumulate gradient steps for this batch over multiple training steps. This strategy allows us to perform efficient sampling while using the PG algorithm. Combined with strategies to filter out samples with small advantage estimates, we call the resulting algorithm Distributed-Aggregated Sampling Handler (DASH). Compared to GRPO, DASH reduces on-policy training time by \textbf{83\%} without sacrificing accuracy (Figure~\ref{fig:time comparison}). We open-source DASH to facilitate further research.

To summarize, our key findings are as follows:
\begin{itemize}
\item For models we consider, we find on-policy RL to be effective but not SFT (Section~\ref{sec:sft_vs_onpolicy}).
\item We propose DASH, and find that it can accelerate on-policy training by \textbf{83\%} without compromising accuracy (Section~\ref{sec:dash_acc}).
\item We find that while PPO-style gradient updates can slightly improve accuracy, it can introduce instability into training (Section~\ref{sec:gradient_update}).
\item We find that removing KL divergence regularization can lead to more concise generations and higher accuracies  (Section~\ref{sec:grpo_com_affect_performance}).
\end{itemize}

\section{Related Work}

\paragraph{LM Reasoning.} Given the promising performance of language models (LMs), numerous studies have explored their application to mathematical problem solving~\citep{hendrycks2021measuringmathematicalproblemsolving, cobbe2021trainingverifierssolvemath, glazer2024frontiermathbenchmarkevaluatingadvanced}, program synthesis~\citep{austin2021programsynthesislargelanguage, puri2021codenetlargescaleaicode}, and other reasoning tasks. Since LMs often exhibit varying performance when directly prompted for these tasks, various methods have been proposed to explicitly elicit reasoning. For instance, Chain-of-Thought prompting~\citep{wei2023chainofthoughtpromptingelicitsreasoning} encourages LMs to generate intermediate reasoning steps before producing the final answer. Tree-of-Thought~\citep{yao2023treethoughtsdeliberateproblem} and Graph-of-Thought~\citep{Besta_2024} extend this idea by imposing logical structure to organize the reasoning process. LM reasoning has also been enhanced through tool use~\citep{yao2023reactsynergizingreasoningacting, shinn2023reflexionlanguageagentsverbal}. While these methods have proven effective in guiding LM reasoning and improving downstream task performance, they primarily focus on better prompt design rather than improving the models’ inherent reasoning capabilities.

\paragraph{RL for LM reasoning.}

Recent efforts have focused on using RL to improve LM reasoning capabilities. In question-answering tasks, FireAct~\citep{chen2023fireactlanguageagentfinetuning} and AgentTuning~\citep{zeng2023agenttuningenablinggeneralizedagent} enhance reasoning capabilities by learning from demonstrations from humans or stronger models. These approaches are commonly referred to as \textit{supervised fine-tuning} (SFT), or \textit{behavior cloning} in the RL literature. However, several studies have found limits on the effectiveness of SFT, instead proposing to use on-policy RL~\citep{deepseekai2025deepseekr1incentivizingreasoningcapability,shao2024deepseekmathpushinglimitsmathematical,zeng2025simplerlzooinvestigatingtamingzero}.

On the other hand, on-policy RL can be very computationally expensive, leading to a great deal of interest in improving efficiency. One shortcoming is that they require re-sampling generations after each model update, leading to sample inefficiency and prolonged training times. To mitigate this, \citet{deepseekai2025deepseekv3technicalreport} propose more efficient transformer architectures to accelerate pretraining, and \citet{kwon2023efficientmemorymanagementlarge} introduce advanced memory management techniques to speed up sampling in post-training. Although current RL algorithms can leverage vLLM acceleration, the full potential of vLLM remains underutilized, leaving significant room for improving RL efficiency.

\section{Effective RL for LM Reasoning}

\begin{figure}
\centering
\includegraphics[width=0.48\textwidth]{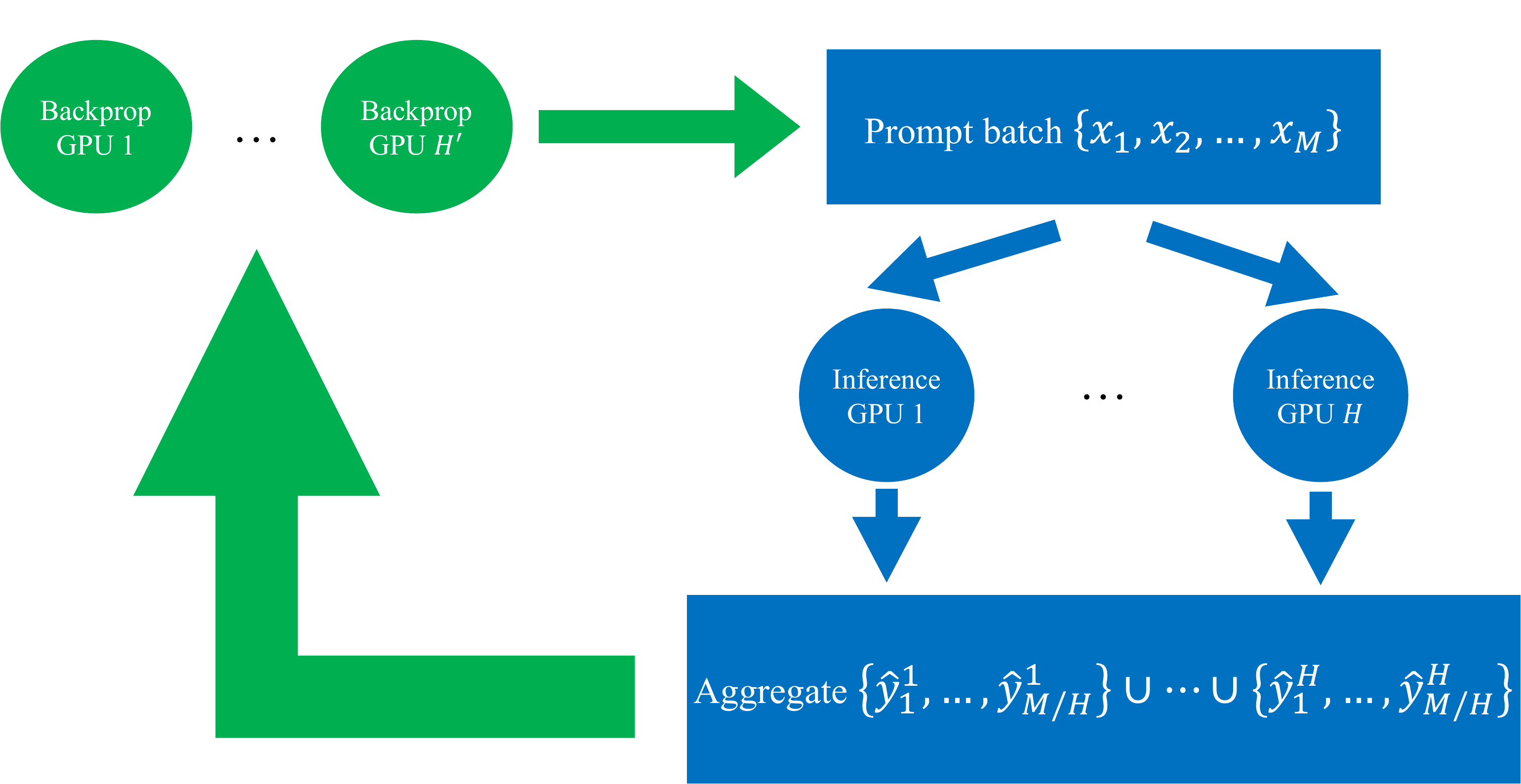}
\caption{Illustration of preemptive sampling. We use $H$ GPUs for inference and $H'$ for backpropagation; they are shown in blue and green, respectively. Given a batch of $M$ prompts $\{\bx_1, \ldots, \bx_M\}$. The inference GPUs then generate corresponding responses $\{\hat{\by}_1, \ldots, \hat{\by}_M\}$, which are aggregated across GPUs into CPU memory. When a backpropagation GPU requests generations for a prompt $\bx_m$, the corresponding cached response $\by_m$ is retrieved and delivered. Since we are using groups for advantage estimation, each prompt $\bx_m$ is duplicated to form groups, and all generations in the same group are sent to the backpropagation GPU upon request.}
\label{fig:preemptive_sampling}
\end{figure}

First, we describe basic design decisions of our RL algorithm rooted in the prior literature; these are based either on experiments from prior work or our own experiments. Specifically, we consider an LM $\pi_{\theta}$ with parameters $\theta$, which takes in a user prompt $\bx$ and generates a reasoning trace $\hat \by$, which we call a \emph{trajectory}. We let $\hat y_t$ denote the $t$th token in trajectory $\hat \by$. For a training prompt $\bx_n$, we can check whether a generated trajectory $\hat\by_n$ produces the correct answer, represented as a scalar reward $r_n=R(\bx_n,\hat\by_n) \in \mathbb{R}$. We assume that $r_n$ is for the entire trajectory; typically, it is a binary indicator of whether the final answer is correct.\footnote{Recent work has found that process rewards~\citep{wang2024mathshepherdverifyreinforcellms} may not be effective in our setting due to the difficulty predicting whether a reasoning trace is on the right track~\citep{deepseekai2025deepseekr1incentivizingreasoningcapability}.}

\subsection{RL Strategy}

The first decision is what kind of RL strategy to use. We consider two strategies: supervised finetuning (SFT) and on-policy RL. SFT is effectively the same as behavior cloning, a popular imitation learning algorithm. Given a prompt training set $D=\{\bx_n\}_{n=1}^N$, SFT collects corresponding expert trajectories $\mathcal{Y}=\{\by_n\}_{n=1}^N$, either from a human or a stronger LM. Then, the LM is optimized via maximizing the log likelihood on $(D,\mathcal{Y})$:
\begin{align*}
\theta^* &= \arg \max_{\theta}  \sum_{n=1}^N \log \pi_{\theta}(\by_n \mid \bx_n) \\
\pi_{\theta}(\by_n\mid\bx_n)&=\prod_{t=1}^T\pi_{\theta}(\hat{y}_{n,t}\mid\bx_n,\hat{y}_{n,1},...,\hat{y}_{n,t-1})
\label{eq:sft}
\end{align*}
Alternatively, on-policy RL learns from trajectories generated by the current LM $\pi_{\theta}$. Given the a prompt set $D$, a typical on-policy RL algorithm optimizes the expected reward:
\begin{align}
\pi_{\theta^*} &= \arg \max_{\theta} J(\theta) \\
J(\theta) &= \frac{1}{N} \sum_{x_n \in D} \mathbb{E}_{\hat\by_n\sim\pi_{\theta}(\cdot\mid\bx_n)}[R(\bx_n,\hat\by_n)]. \nonumber
\end{align}
While SFT has been shown to be effective in settings such as~\citet{muennighoff2025s1simpletesttimescaling}, they use a post-trained larger sized model (Qwen2.5-32B-Instruct); our experiments show that it can be ineffective when the gap between the expert and $\pi_{\theta}$ is too large (specifically, we use small base models instead of larger instruction-tuned models). For instance, an expert may take leaps of reasoning that are incomprehensible to the learner. Thus, DASH uses on-policy RL. Another alternative that has been studied in the literature is self-imitation~\citep{oh2018selfimitationlearning}, where the ``expert'' trajectories are obtained by performing search guided by $\pi_{\theta}$, but results applying this strategy to LMs have so far been mixed~\citep{shao2024deepseekmathpushinglimitsmathematical}.

\subsection{Gradient Update Strategy}
\label{sec:PGtoPPOGRPO}

Next, we discuss the gradient update strategy. We consider both policy gradient (PG)~\citep{NIPS1999_464d828b} and PPO~\citep{schulman2017proximalpolicyoptimizationalgorithms} (which includes GRPO~\citep{shao2024deepseekmathpushinglimitsmathematical}). In general, we consider gradient approximations $\nabla_{\theta} J(\theta)\approx N^{-1} \sum_{n=1}^NJ_n$ where $J_n$ encodes the gradient approximation for the $n$th summand of $J(\theta)$. First, by the Policy Gradient Theorem, using
\begin{align}
J_n^{\text{PG}}=\mathbb{E}_{\by_n\sim\pi_{\theta}(\cdot\mid\bx_n)}\left[\frac{\nabla_{\theta} \pi_{\theta}(\hat \by_n\mid \bx_n)}{\pi_{\theta}(\hat \by_n\mid \bx_n)} A^{\pi_{\theta}}(\bx_n, \hat \by_n)\right]
\label{eq:pg}
\end{align}
is exact, i.e., $\nabla_{\theta}J(\theta)=N^{-1}\sum_{n=1}^NJ_n^{\text{PG}}$.
Here, $A^{\pi_{\theta}}(\bx_n,\hat\by_n)$ is the \emph{advantage function}, which we discuss below. This update is truly on-policy since the trajectories $\hat\by$ must be sampled using the current policy $\pi_{\theta}$. In robotics, PG can be unstable due to high variance when estimating the gradient $\nabla_{\theta} \pi_{\theta}(\hat \by_n\mid \bx_n)$; as a consequence, $\pi_{\theta}$ can change rapidly across gradient steps, sometimes even becoming worse. PPO was devised to mitigate this instability. Specifically, they weaken the on-policy requirement, and ``freeze'' the data-generating policy $\pi_{\theta_{\text{old}}}$ for some number of gradient steps. The resulting update has the alternative form
\begin{align*}
&J_n^{\text{PPO}} = \\
&\mathbb{E}_{\hat\by_n\sim\textcolor{red}{\pi_{\theta_{\text{old}}}}(\cdot\mid\bx_n)}\left[\frac{\nabla_{\theta} \pi_{\theta}(\hat \by_n\mid \bx_n)}{\textcolor{red}{\pi_{\theta_{\text{old}}}}(\hat \by_n\mid \bx_n)} A^{\textcolor{red}{\pi_{\theta_{\text{old}}}}}(\bx_n, \hat \by_n)\right],
\end{align*}
where the differences compared to \eqref{eq:pg} are highlighted in red. Because this gradient is only valid when $\theta\approx\theta_{\text{old}}$, a KL regularization is imposed, to obtain $J_n^{\text{PPO-KL}} = J_n^{\text{PPO}} + \beta J_n^{\text{KL}}$, where
\begin{align*}
J_n^{\text{KL}} &= \nabla_{\theta} D_{\text{KL}}(\pi_{\theta_{\text{base}}}(\cdot\mid\bx_n)\;\|\;\pi_{\theta}(\cdot\mid\bx_n))
\end{align*}
Following \citet{jaques2019wayoffpolicybatchdeep,ouyang2022training}, the KL divergence term is with respect to the original model $\pi_{\theta_{\text{base}}}$ instead of $\pi_{\theta_{\text{old}}}$ as in PPO. 

Critically, in PPO, $\theta_{\text{old}}$ is updated to be $\theta$ every $K$ steps, where $K$ is a hyperparameter. To further improve stability, the gradient is often clipped. GRPO uses the same gradient update as PPO; early versions include a weight $1/\text{len}(\hat\by_n)$ on the $n$th term to normalize by the length of the trajectory, but this term was removed in later versions~\citep{liu2025understandingr1zeroliketrainingcritical}. Finally, we note that when $\theta=\theta_{\text{old}}$, this gradient update is equivalent to the PG update (\ref{eq:pg}); this property holds even with gradient clipping.

Now, assume we have sampled a batch of $M$ samples $\{(\bx_m, \hat \by_m)\}_{m=1}^M$ from $\pi_{\theta_{\text{old}}}$, where initially $\theta=\theta_{\text{old}}$. If we take a single gradient step on all examples, then PPO coincides with PG. This is the strategy used by DASH. We consider two implementations of PPO that do not devolve into PG. First, we can take $K$ gradient steps using all $M$ samples, which we call \emph{PPO-Multi} (or just \emph{Multi}). Second, we can divide the $M$ examples into $K$ mini-batches of size $M/K$ each, and take one gradient step on each mini-batch, which we call \emph{PPO-Mini} (or just \emph{Mini}). In our experiments, we find that DASH is more stable than Multi and Mini, suggesting that the added complexity of PPO-based off-policy gradient updates increases variance.

We include the KL term in DASH (i.e., $J_n^{\text{DASH}}=J_n^{\text{PG}}+\beta J_n^{\text{KL}}$) for closer comparison of DASH to Multi and Mini. However, in our experiments, we find that omitting the KL term improves accuracy.

\subsection{Advantage Estimation}

A key challenge in RL is estimating the quantity $A^{\pi}(\hat\by\mid\bx)$, which is called the \emph{advantage}~\citep{10.5555/3312046}; it is defined to be $A^{\pi}(\hat\by\mid\bx)=Q^{\pi}(\hat\by\mid\bx)-V^{\pi}(\bx)$, where $Q^{\pi}$ is the Q-function and $V^{\pi}$ is the value function. Intuitively, it captures how the specific generation $\hat\by$ compares to a random sample $\hat\by'\sim\pi_{\theta}(\cdot\mid\bx)$. In general, $A^{\pi}$ is not known and must be estimated from data. We consider three strategies: (i) training a model to predict $A^{\pi}$, (ii) a Monte Carlo estimate called the \emph{single-path method}, and (iii) a Monte Carlo estimate introduced by GRPO. The first approach is to train a model to predict $Q^{\pi}(\hat\by\mid\bx)$, which can be used to compute $V^{\pi}$ and $A^{\pi}$~\citep{schulman2017trustregionpolicyoptimization}. This approach can reduce variance, but recent work has found that it is highly biased due to the difficulty in predicting $Q^{\pi}$ for reasoning tasks~\citep{liang2022reducingvariancetemporaldifferencevalue}. Thus, we focus on Monte Carlo approaches.

The most popular Monte Carlo approach is the \emph{single-path method}, which uses the estimate
\begin{align}
\label{eqn:singlepath}
A^{\pi_{\theta}}(\hat\by_n\mid\bx_n)\approx r_n-
\frac{1}{N}\sum_{n'=1}^Nr_{n'},
\end{align}
i.e., it is the centered reward; $b=N^{-1}\sum_{n'=1}^Nr_{n'}$ is called the \emph{baseline}. Intuitively, $r_n$ is an estimate of the Q-function, and $b$ is an estimate of the value function. A standard modification is to normalize by the standard deviation; this normalization can be useful when rewards tend to increase significantly as learning progresses, but our rewards are bounded so this cannot happen. Another modification is to leave out the reward for rollout $n$ when estimating the value for rollout $n$, which reduces bias~\citep{10.5555/3312046}; this modification can be important when $N$ is small (e.g., $N=2$) but only has a minor impact for larger $N$ since the bias is small.

A shortcoming of the single-path method is that $b$ is an estimate of the average value $N^{-1}\sum_{n'=1}^NV(\bx_{n'})$ across all samples, whereas it ideally should estimate the value $V(\bx_n)$. One alternative is the \emph{vine method}~\citep{kazemnejad2024vineppounlockingrlpotential, schulman2017trustregionpolicyoptimization}, which uses a targeted sampling strategy to fix this issue; however, the vine method requires a large number of samples, making it computationally expensive. GRPO uses an advantage estimate that interpolates between the single-path and vine methods. It exploits the fact that in the reasoning setting, we typically train on multiple samples $\hat\by_n$ for a single user prompt $\bx_n$. In our formulation, we can think of there being multiple $\bx_n$ that are identical. Suppose that we partition $N$ into groups $N_1,...,N_K$, where $\bx_n$ is the same for all $n\in N_k$. Then, it estimates the advantage using the formula
\begin{align}
\label{eq:grpoadvantage}
A^{\pi_{\theta}}(\hat\by_n\mid\bx_n)\approx r_n-\frac{1}{N_k}\sum_{n'\in N_k}r_{n'},
\end{align}
where $N_k$ is the group containing $n$.  In other words, it replaces the baseline with a state-dependent baseline $b(\bx_n)=N_k^{-1}\sum_{n'\in N_k}r_{n'}$; now, $b(\bx_n)$ is an unbiased estimate of $V(\bx_n)$. This strategy can be viewed as performing a vine estimate of the advantage at state $\bx_n$, but not at any other state. DASH uses the GRPO advantage estimate (\ref{eq:grpoadvantage}).

\subsection{Preemptive Sampling}
\label{sec:preemptive_sampling}

A key feature of RL for LMs is that inference typically occurs on specialized inference servers such as vLLM~\citep{kwon2023efficient}. Importantly, inference is typically much more memory efficient than backpropagation, meaning much larger batches are optimal for inference compared to backpropagation. Empirically, sampling takes up a much larger portion of training time than backpropagation if performed in small batches (Figure~\ref{fig:time comparison}). Thus, we propose \emph{preemptive sampling}, where we sample a large number of trajectories in one batch, and then perform backpropagation on these samples in smaller batches. Preemptive sampling can be further sped up by using multiple inference servers in parallel (Figure~\ref{fig:preemptive_sampling}). In practice, our method can be used for both on-policy and off-policy sampling, depending on algorithmic design choices, as detailed in Section~\ref{sec:PGtoPPOGRPO}. Figure~\ref{fig:preemptive_sampling} illustrates preemptive sampling. DASH uses preemptive sampling.

\subsection{Gradient Filtering}
\label{sec:gf}

Finally, we propose to drop examples with small advantage estimates (which is equivalent to clipping small advantage values to zero, effectively dropping them from the gradient update). If the advantage estimate is small, then the contribution to the gradient is likely to be small (unless $\nabla_{\theta}\pi_{\theta}(\hat\by_n\mid\bx_n)$ happens to be very large, which we find to be unlikely in practice). Intuitively, these are examples where the model either almost always gets the answer right (in which case there is nothing new to learn) or almost always gets it wrong (in which case the problem is currently too difficult to learn). In addition, even if we only drop advantages that are identically zero, this strategy can provide a speedup since backpropagation still takes time to compute the gradients $\nabla_{\theta}\pi_{\theta}(\hat\by_n\mid\bx_n)$ before they are eventually multiplied by $A^{\pi_{\theta}}(\hat\by_n\mid\bx_n)=0$. DASH uses gradient filtering.

\section{Experimental Results}

We perform experiments showing that (i) on-policy RL significantly outperforms SFT (Section~\ref{sec:sft_vs_onpolicy}, (ii) DASH significantly reduces running time compared to standard GRPO (Section~\ref{sec:dash_acc}), (iii) PG gradient updates outperform PPO-based gradient updates (Section~\ref{sec:gradient_update}), and (iv) removing KL divergence can lead to more concise generations and higher accuracies (Section~\ref{sec:grpo_com_affect_performance}).

\subsection{Experimental Setup}

We use Qwen2.5-\{0.5B, 1.5B, 3B\} models as our base models, all of which are not post-trained (i.e., no instruction tuning). We use the MATH dataset~\citep{hendrycks2021measuringmathematicalproblemsolving}, with the MATH-500 split ~\citep{lightman2023letsverifystepstep}, which contains 12,000 examples for training and 500 examples for evaluation.
We additionally use the GSM8K dataset~\citep{cobbe2021trainingverifierssolvemath} for out-of-distribution evaluation, which contains 1,319 examples. Finally, we also perform some experiments in the coding domain using the MBPP+ dataset~\citep{evalplus}, a 378-problem subset of verified problems from the MBPP dataset~\citep{austin2021programsynthesislargelanguage}; we use 264 problems for training and 114 for evaluation. Additional details are provided in Table~\ref{tab:qwen2p5_hparams}.

\subsection{SFT vs. On-Policy RL}
\label{sec:sft_vs_onpolicy}

We compare three algorithms: (i) SFT with human-written reasoning traces, denoted \textit{SFT-H}, (ii) SFT with reasoning traces from Qwen2.5-7B-Instruct, denoted \textit{SFT-M}, and (iii) DASH. Results are shown in Table~\ref{tab:sft_onpolicy}. As can be seen, DASH improves performance both in-distribution and out-of-distribution, demonstrating that on-policy algorithms can efficiently learn mathematical reasoning skills that generalize across datasets. On the other hand, neither SFT-H nor SFT-M improve performance, with SFT-H significantly degrading both in-distribution and out-of-distribution performance. Intuitively, the substantial performance degradation caused by SFT-H can be attributed to the fact that human reasoning often omits many intermediate steps, which is especially problematic for smaller LMs.

For coding, we train on human programs in MBPP+. Results are shown in Table~\ref{tab:code_onpolicy}.
As can be seen, DASH outperforms SFT in most cases, demonstrating the the general effectiveness of on-policy RL at improving the reasoning capabilities of LMs. To the best of our knowledge, these are among the first results to show that on-policy RL can improve code generation for smaller LMs.

\begin{table}[t]
\centering
\begin{tabular}{lccc}
\toprule
\multicolumn{1}{c}{Method} & Size (B) & MATH (\%) & GSM8K (\%) \\
\midrule
\multicolumn{1}{c}{Base} & 0.5 & $22.6$     & $30.3$      \\
& 1.5 & $48.0$     &  $58.8$     \\
& 3.0   & $58.8$     & $66.0$      \\ \midrule
\multicolumn{1}{c}{SFT-H} & 0.5 & $8.0$     & $7.2$      \\
& 1.5 & $17.2$     & $32.8$      \\
& 3.0   & $24.0$     & $30.6$      \\ \midrule
SFT-M                     & 0.5 & $24.0$     & $22.7$      \\
& 1.5 & $46.2$     & $46.0$      \\
& 3.0   & $53.0$     & $66.0$      \\ \midrule
DASH                      & 0.5 & $27.2$    & $31.1$      \\
& 1.5 & $54.0$     & $58.8$      \\
& 3.0   & $64.6$     & $64.6$      \\ \bottomrule
\end{tabular}
\caption{Comparison of SFT to on-policy RL on math.}
\label{tab:sft_onpolicy}
\end{table}

\begin{table}[t]
\centering
\begin{tabular}{lccc}
\toprule
\multicolumn{1}{c}{Method}                       & Size (B) & pass\@1 (\%) & pass@8 (\%) \\
\midrule
\multicolumn{1}{c}{BASE} & 0.5 & $2.6$     & $22.8$      \\
& 1.5 & $7.1$     &  $60.5$     \\
\midrule
\multicolumn{1}{c}{SFT-H} & 0.5 & $8.77$     & $29.0$      \\
& 1.5 & $19.3$     &  $42.1$     \\\midrule
DASH                      & 0.5 & $11.4$    & $40.4$      \\
& 1.5 & $23.7$     & $63.2$      \\\bottomrule
\end{tabular}
\caption{Comparison of SFT to on-policy RL on coding.}
\label{tab:code_onpolicy}
\end{table}

\begin{table}[t]
\centering
\begin{tabular}{cccc}\toprule
Method & Time (h) & MATH (\%) & GSM8K (\%)\\ \midrule
BASE & N/A & 22.6  & 30.3 \\ 
GRPO   &  38.9        & $27.6$       &  $32.8$ \\
No-GF   &   6.9       &  $27.4$    & $31.6$    \\ 
DASH   &   6.6       &  $27.2$    & $31.1$    \\ \bottomrule
\end{tabular}
\caption{Comparing on-policy RL algorithms on Qwen2.5-0.5B for math.}
\label{tab:acceleration}
\end{table}

\begin{table}[t]
\centering
\begin{tabular}{cccc}\toprule
Method & Time (m) & pass@1 (\%) & pass@8 (\%)\\ \midrule
BASE & N/A & 2.3  & 22.8 \\ 
GRPO   &  35.3        & $11.4$       &  $49.1$ \\
No-GF   & 16.3         &  $10.5$    & $43.9$    \\ 
DASH   &   16.5       &  $11.4$    & $40.4$    \\ \bottomrule
\end{tabular}
\caption{Comparing on-policy RL algorithms using Qwen2.5-0.5B for coding.}
\label{tab:acceleration_coding}
\end{table}

\subsection{DASH vs. GRPO}
\label{sec:dash_acc}

Next, we compare DASH to GRPO both in terms of accuracy and running time by training Qwen2.5-0.5B using both GRPO and DASH. We also use an ablation of DASH without gradient filtering, denoted \textit{No-GF}. Results are shown in Table~\ref{tab:acceleration} and illustrated in Figure~\ref{fig:time comparison}. As can be seen, DASH significantly reduces GRPO training time (from 39 hours to 6.6 hours) without any significant reduction in performance, highlighting the effectiveness of preemptive sampling and gradient filtering.

Compared to No-GF, DASH reduces running time by 4\% without any significant reduction in performance. The effectiveness of gradient filtering can be improved; see Appendix~\ref{sec:additional_experiment}. We additionally show results for coding in Table~\ref{tab:acceleration_coding}. For coding, we again see a significant speedup, though it is smaller since the generation length is much smaller so the gap in optimal inference and backpropagation batch sizes is smaller.

\begin{figure}[t]
\centering
\begin{subfigure}[b]{0.4\textwidth}
\centering
\includegraphics[width=\textwidth]{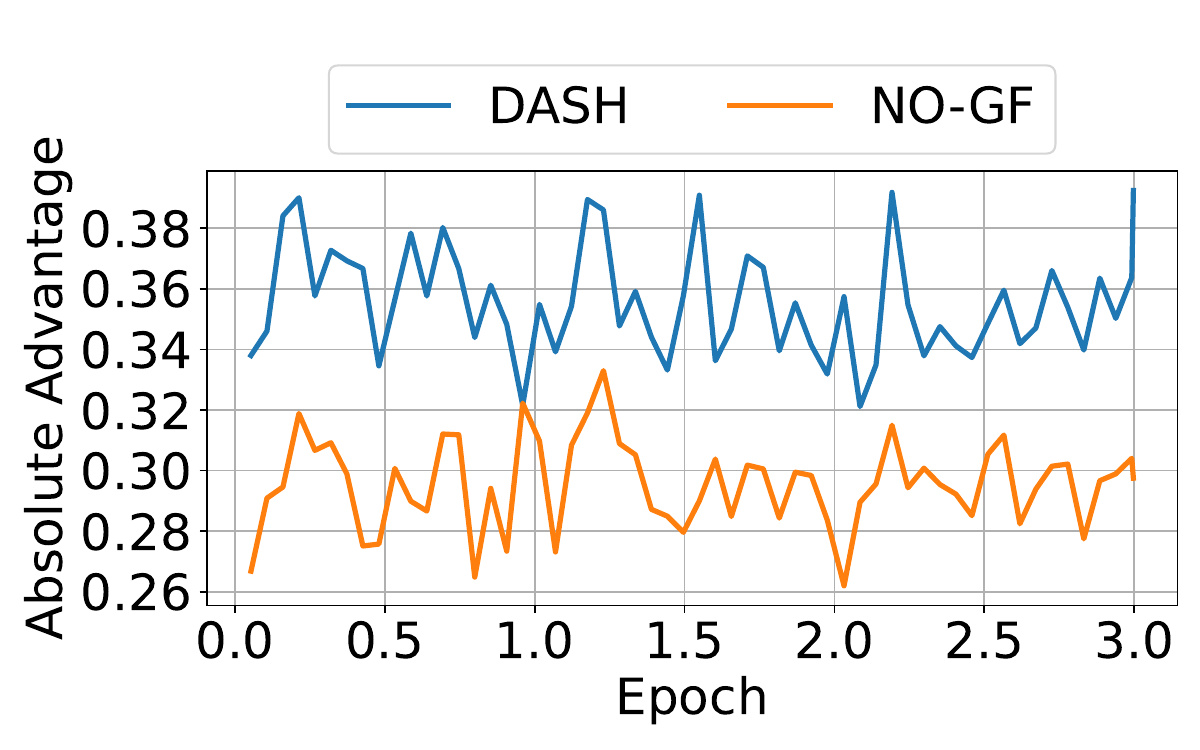}
\caption{Average absolute advantage values.}
\label{fig:mean_advantage}
\end{subfigure}
\
\begin{subfigure}[b]{0.4\textwidth}
\centering
\includegraphics[width=\textwidth]{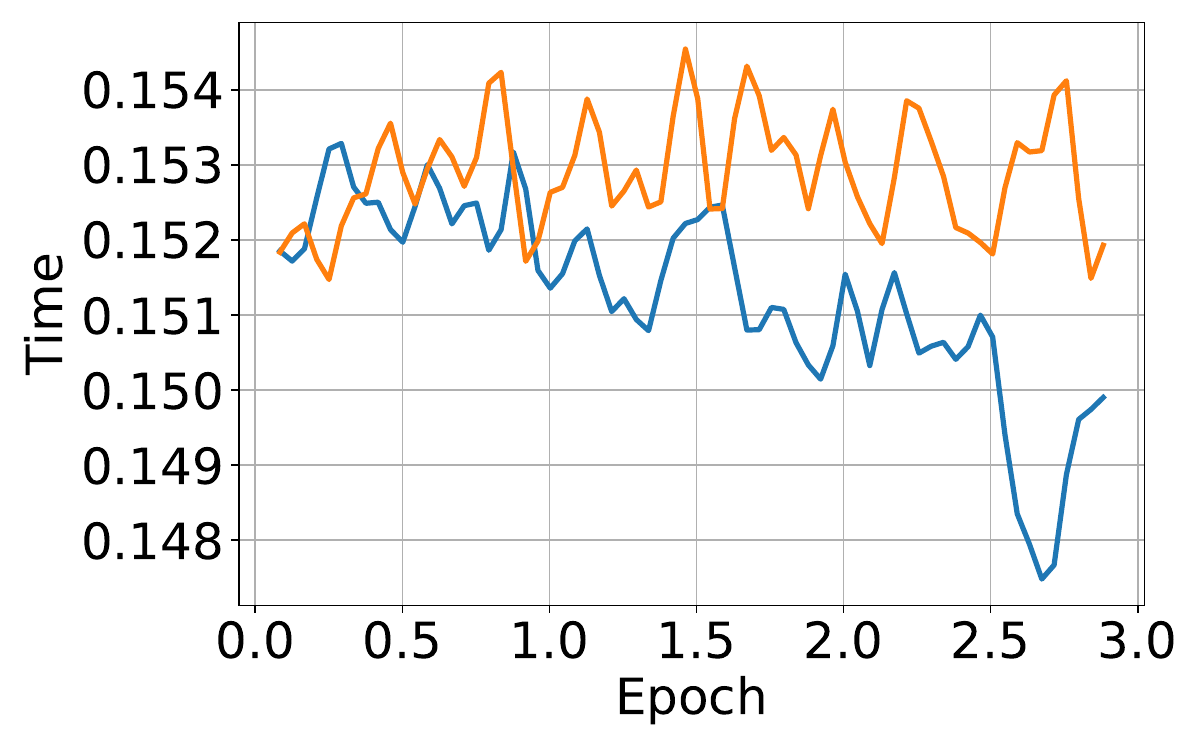}
\caption{Loss computation time.}
\label{fig:loss_computation_time}
\end{subfigure}
\
\begin{subfigure}[b]{0.4\textwidth}
\centering
\includegraphics[width=\textwidth]{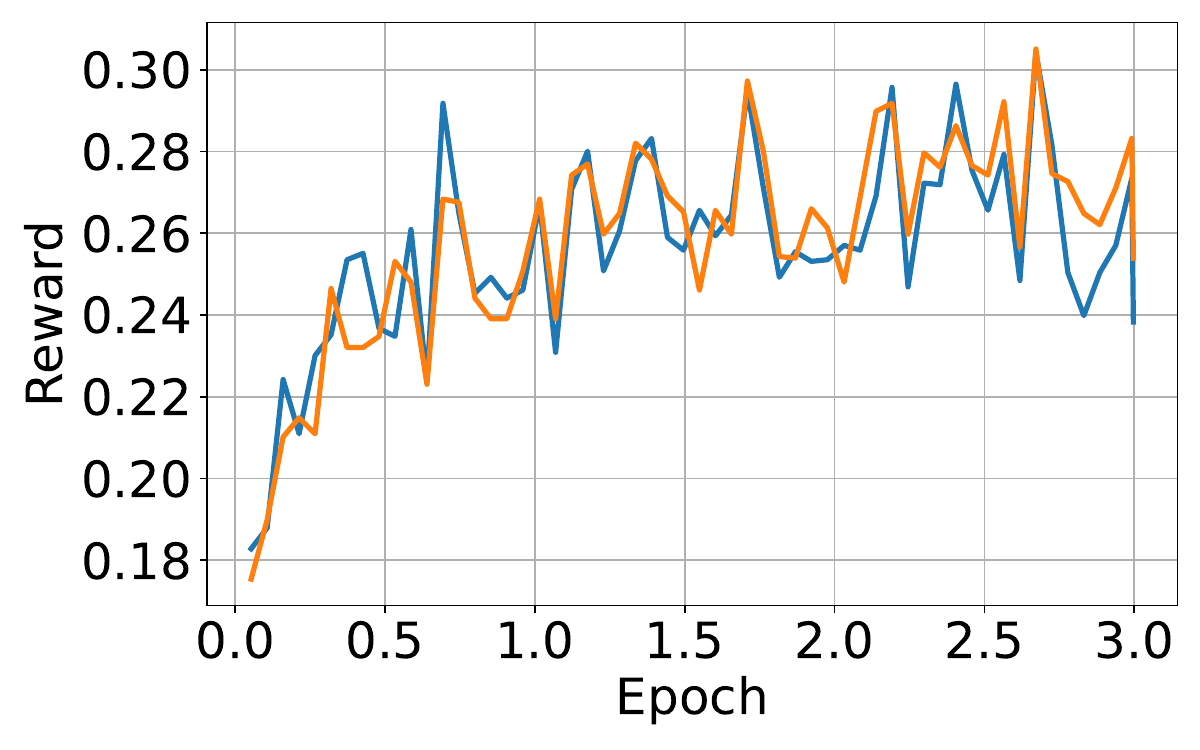}
\caption{Training reward recurves.}
\label{fig:reward_gf}
\end{subfigure}
\caption{Comparison between DASH and No-GF for Qwen2.5-0.5B on math.}
\label{fig:three graphs}
\end{figure}

The impact of GF on training dynamics is illustrated in Figure~\ref{fig:three graphs}. Specifically, as shown in Figure~\ref{fig:mean_advantage}, gradient filtering increases the average absolute advantage values, leading to more significant gradient updates; consequently, as shown in Figure~\ref{fig:loss_computation_time}, forward and backward pass running times are reduced. Finally, since only samples inducing trivial gradient updates are filtered out, the training curves remain similar before and after applying gradient filtering, as shown in Figure~\ref{fig:reward_gf}.

\subsection{PG vs. PPO Gradient Updates}
\label{sec:gradient_update}

\begin{table}[t]
\centering
\begin{tabular}{lccc}
\toprule
\multicolumn{1}{c}{Method}                       & Size (B) & MATH (\%) & GSM8K (\%) \\ \midrule
\multicolumn{1}{c}{DASH} & 0.5 & 27.2     & 31.1      \\
& 1.5 & 54.0     &  58.8     \\\midrule
\multicolumn{1}{c}{Multi} & 0.5 & $28.8$     & $31.5$      \\
& 1.5 & 54.0     & 61.0      \\\midrule
\multicolumn{1}{c}{Mini} & 0.5 & $29.8$     & $31.6$       \\
& 1.5 & 36.0     & 8.1      \\
\bottomrule
\end{tabular}
\caption{Comparing PG to PPO for math.}
\label{tab:grpo_comp}
\end{table}

\begin{figure}[t]
\centering
\includegraphics[width=\linewidth]{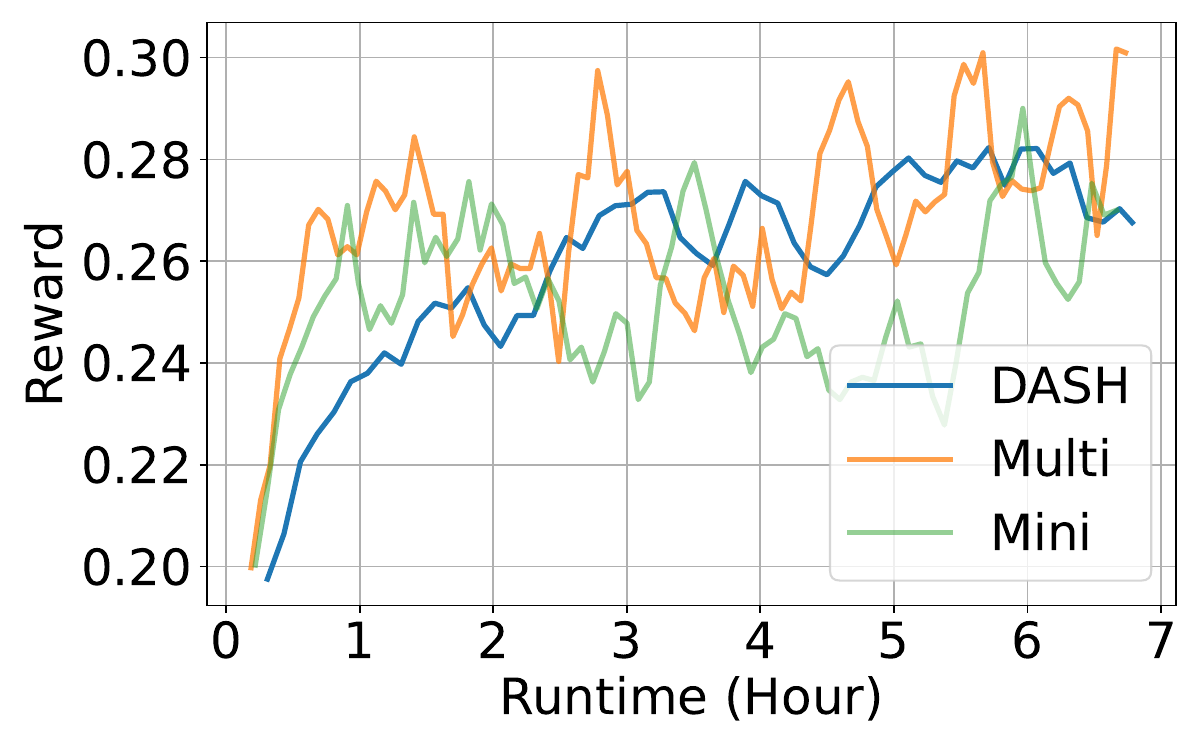}
\caption{Training reward curves for PG vs. PPO on Qwen2.5-0.5B for math.}
\label{fig:reward_curves}
\end{figure}

Next, we compare DASH to Multi and Mini.
DASH uses a batch size of $M=256$ (with $K=1$), Multi uses $M=256$ and $K=3$, and Mini uses $M=8$ so $K=32$.
Multi and Mini are slower than DASH; for fair comparison, we truncate their training times to match the wall-clock time of DASH. The results are shown in Table~\ref{tab:grpo_comp}, and training curves are shown in Figure~\ref{fig:reward_curves}. As can be seen, Multi and Mini achieve faster initial performance improvements and have slightly higher accuracies; however, they have significantly more unstable training curves. Similar results for the 1.5B model are shown in Appendix~\ref{sec:additional_experiment}.

\subsection{KL Divergence Regularization}
\label{sec:grpo_com_affect_performance}

\begin{figure}[t]
\centering
\begin{subfigure}[b]{0.4\textwidth}
\centering
\includegraphics[width=\linewidth]{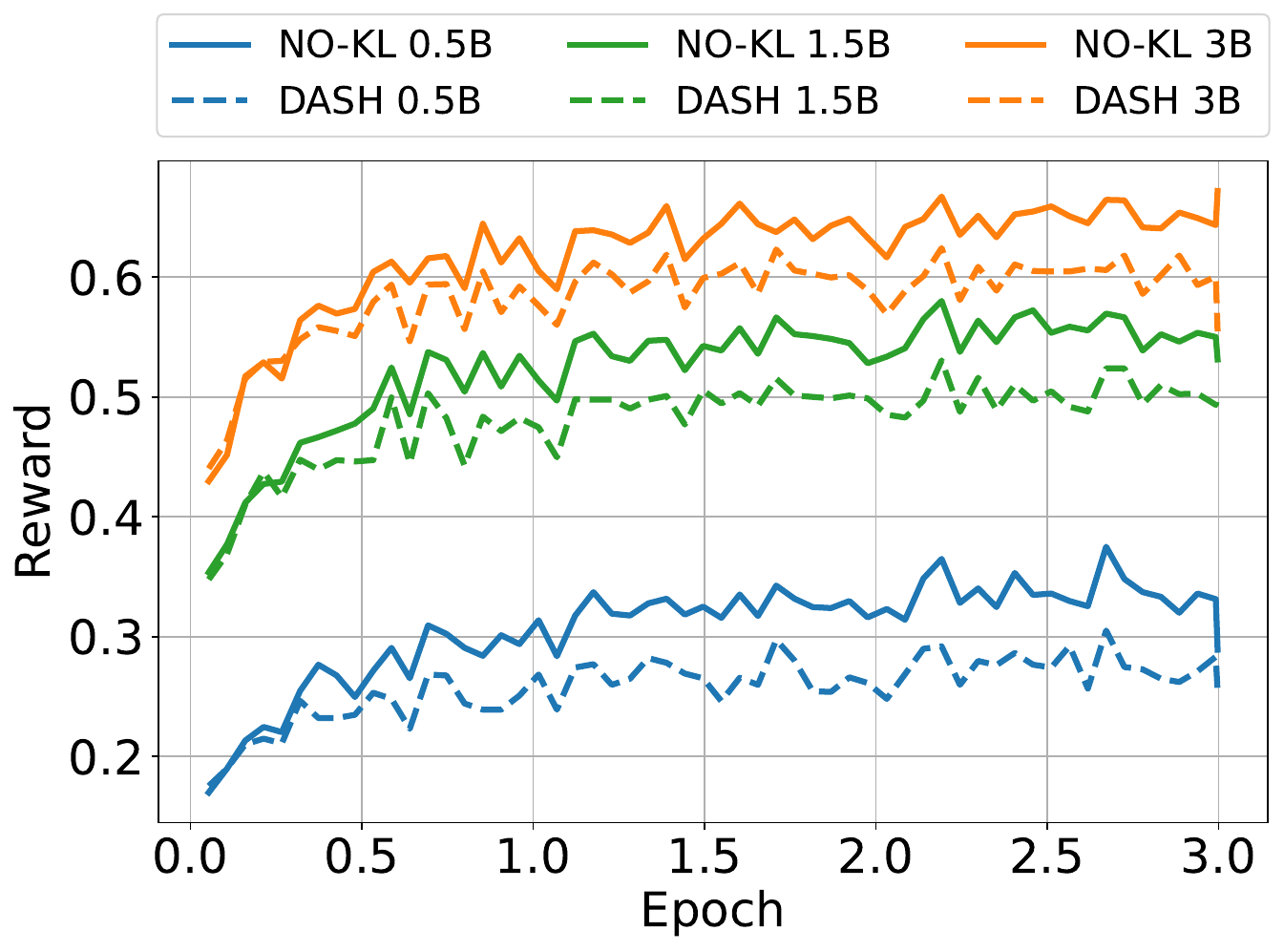}
\caption{Training reward curves.}
\label{fig:no_kl}
\end{subfigure}
\
\begin{subfigure}[b]{0.4\textwidth}
\centering
\includegraphics[width=\textwidth]{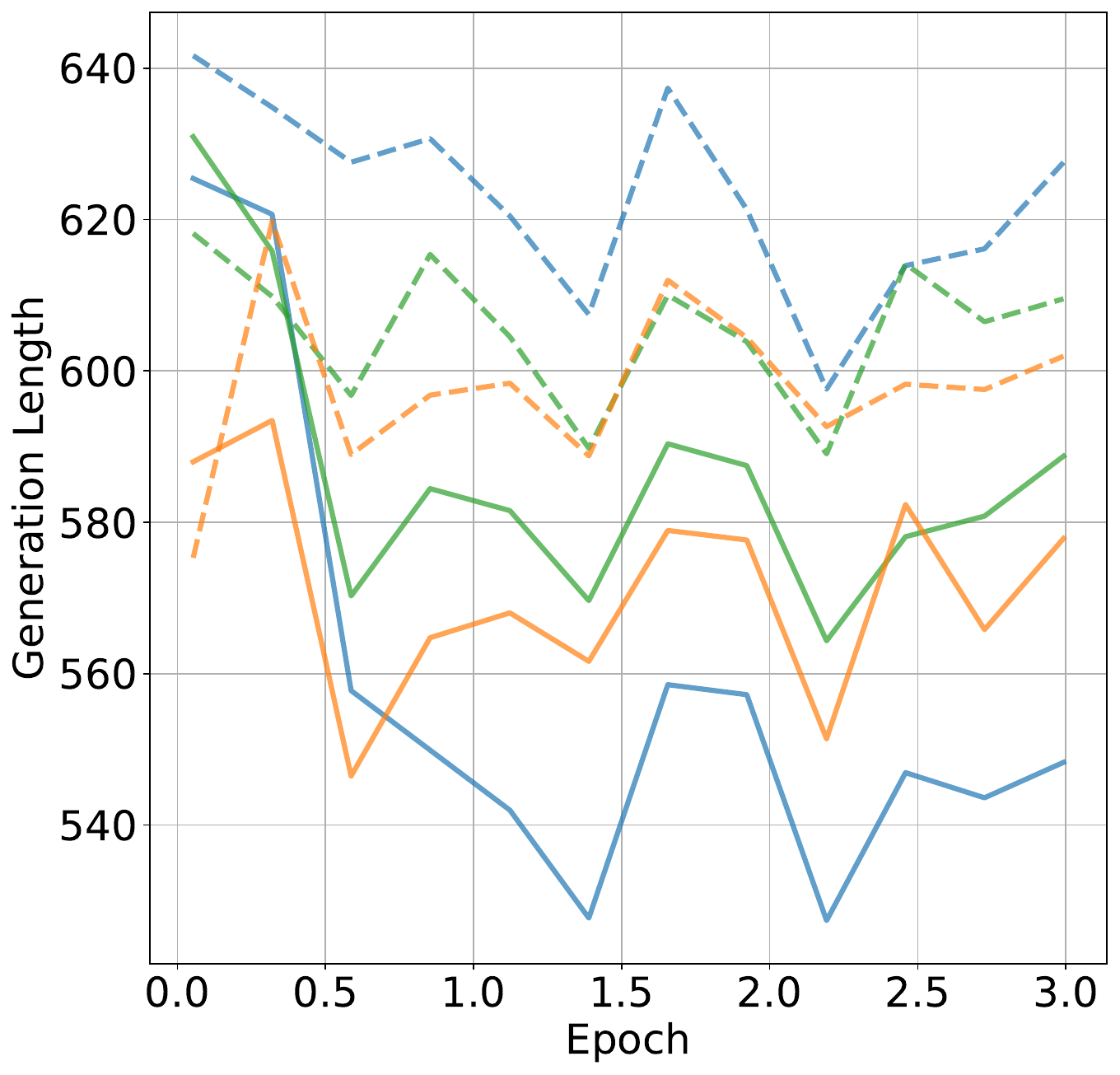}
\caption{Generation length.}
\label{fig:no_kl_length}
\end{subfigure}
\caption{Comparing KL divergence regularization on math.}
\label{fig:no_kl_comparison}
\end{figure}

Next, we compare DASH to an ablation without the KL divergence term, denoted \textit{No-KL}. Training reward curves are shown in Figure~\ref{fig:no_kl}. As can be seen, removing KL divergence regularization generally leads to higher rewards during training; most likely, No-KL can focus on reward optimization without being constrained to stay close to the initial model. As shown in Table~\ref{tab:no_kl}, No-KL achieves greater in- and out-of-distribution accuracy than DASH (except in the case of the out-of-distribution accuracy of the 3B model).

Furthermore, as shown in Figure~\ref{fig:no_kl_length}, we find that for No-KL, the average generation length is shorter, thereby reducing overall training time; this difference is also reflected in Table~\ref{tab:no_kl}. We hypothesize that to compensate for KL divergence regularization, models must generate longer reasoning traces.

\begin{figure}[t]
\centering
\includegraphics[width=0.8\linewidth]{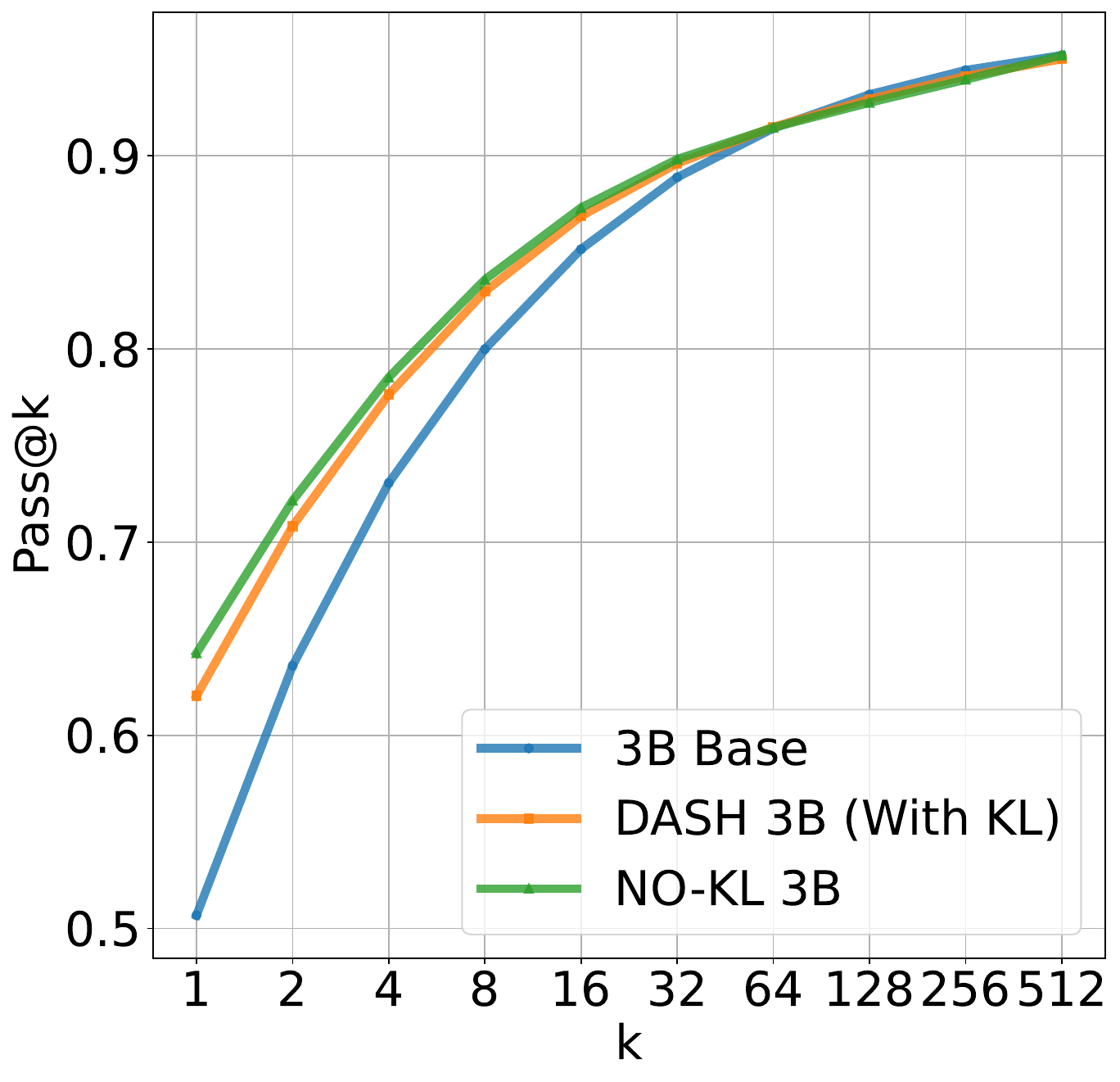}
\caption{Impact of KL divergence regularization on pass@k for Qwen2.5-3B on math.}
\label{fig:pass}
\end{figure}

Finally, for the 3B model, we study how KL divergence regularization affects pass@k. We follow \citet{chen2021evaluatinglargelanguagemodels} to evaluate pass@k in an unbiased way. Results are in Figure~\ref{fig:pass}: No-KL performs best for small $k$, although the gap closes for larger $k$. Intuitively, RL concentrates probability mass and reduces generation diversity~\citep{shypula2025evaluatingdiversityqualityllm, west2025basemodelsbeataligned}.

\begin{table}[t]
\adjustbox{max width=\linewidth}{%
\centering
\begin{tabular}{lcccc}
\toprule
\multicolumn{1}{c}{Method} & Size (B) & Time (h) & MATH (\%) & GSM8K (\%) \\ \midrule
& 0.5 & N/A & $22.6$     & $30.3$      \\
\multicolumn{1}{c}{Base} & 1.5 & N/A & $48.0$     &  $58.8$     \\
& 3.0 & N/A & $58.8$     & $66.0$      \\ \midrule
 & 0.5 & 6.6 & $27.2$ & 31.1       \\
\multicolumn{1}{c}{DASH (with KL)} &1.5 & 12.8 & 54.0     & 58.8      \\
&3.0 & 22.6 & 64.6    & 64.6     \\ \midrule
& 0.5 & 5.7 & $31.4$ & $34.0$       \\
\multicolumn{1}{c}{No-KL} &1.5 & 10.3 & 56.8     & 62.1      \\
&3.0 & 17.6& 66.4     & 60.0      \\ \bottomrule
\end{tabular}}
\caption{Comparing No-KL to DASH and Base on math.}
\label{tab:no_kl}
\end{table}

\section{Conclusion}
We have performed a careful empirical analysis of several key design decisions in RL algorithms for improving language model reasoning, with a focus on computationally constrained scenarios; these include SFT vs. on-policy RL, policy gradient vs. PPO, and whether KL divergence regularization is used. Furthermore, we identify the sampling strategy as the primary computational bottleneck in on-policy RL; to address these issues, we propose DASH, a novel algorithm using preemptive sampling and gradient filtering to improve efficiency. We demonstrate that DASH can reduce RL training time by 83\% while maintaining performance. More broadly, we believe that systematizing the study of RL for language model reasoning is key to designing more effective RL algorithms in this domain, which differs significantly from robotics domains targeted by existing RL algorithms such as PPO. Our study is a first step in this direction.

\paragraph{Limitations.}
Due to computational constraints, we do not perform extensive hyperparameter tuning, instead adopting commonly used values; better hyperparameter choices could further improve performance. Also, we focus on relatively small Qwen2.5 models, and our findings may not generalize to larger models or other architectures.

\paragraph{Ethics statement.}
Our paper aims to design better RL algorithms for reasoning in LMs; we do not foresee any ethical concerns beyond standard ethical issues with reasoning in LMs.

\bibliography{custom}

\clearpage
\appendix

\section{Additional Experimental Setup}
\label{sec: exp_detail}


\begin{table*}[t]
\centering
\small
\begin{tabular}{@{} l c c c l @{}}
\toprule
Hyperparameter & Qwen2.5-0.5B & Qwen2.5-1.5B & Qwen2.5-3B \\
\midrule
NVIDIA A6000 GPUs (training / sampling) & 4 / 2 (2 / 2 using ZeRO) & 4 / 2 & 4 / 2  \\
Learning rate for DASH based runs & $1\times10^{-6}$ & $1\times10^{-6}$ & $1\times10^{-6}$   \\
Learning rate for Multi and Mini & $1\times10^{-6}$ & $3\times10^{-6}$ & N/A   \\
Epochs & 3&3&3\\
Batch size per device & 2 & 2 & 2 \\
Generations per prompt & 4 & 4 & 4   \\
Max completion length (tokens) & 2048 & 2048 & 2048  \\
Gradient accumulation steps for DASH based runs & 32 & 32 & 32 \\
Gradient accumulation steps for Multi and Mini & 32 & 128 & N/A \\
Gradient steps per sampled batch for Multi & 3 & 3 & N/A \\
Gradient‐filtering threshold & 0.1 & 0.1 & 0.1 \\
Normalize gradients by generation length? & No & No & No \\
\bottomrule
\end{tabular}
\caption{Experimental configuration and hyperparameters for on-policy RL on MATH.}
\label{tab:qwen2p5_hparams}
\end{table*}

\paragraph{Math.}
All GRPO experiments are conducted using 6 Nvidia
A6000 GPUs; we use 4 GPUs for backpropagation
and 2 for inference across all three model sizes (Qwen2.5-\{0.5B, 1.5B, 3B\}) (in practice, the 0.5B model only needs 2 GPUs for backpropagation, but we still use 4 for consistency).

Our implementation is based on Huggingface's GRPO Trainer;
the hyperparameters are as follows: 
\begin{itemize}
\item Learning rate: 1e-06; for comparing to Multi and Mini on Qwen2.5-1.5B, we use 3e-06 due to larger batch size
\item Backpropagation batch size per GPU: 2 (so batch size is 8)
\item \# generations per prompt: 4 (resulting in 2 prompts backpropagated on in each step)
\item Maximum completion length: 2048
\item Inference batch size for DASH: 256 (128 per inference GPU); for comparing to Multi and Mini on Qwen2.5-1.5B, we use 1024 (512 per inference GPU)
\item Gradient accumulation steps for DASH: 32 (so the effective batch size is 256); for comparing to Multi and Mini on Qwen2.5-1.5B, we use 128 (so the effective batch size is 1024)
\item Gradient steps per batch for Multi: 3
\item Batch size for Mini: 8 (equivalently, no gradient accumulation)
\item Gradient filtering threshold: 0.1
\end{itemize}
All other parameters are set to the default of the Huggingface trainer; a summary of the GRPO hyperparamaters is in Table~\ref{tab:qwen2p5_hparams}. To reduce memory footprint, we use DeepSpeed~\citep{rajbhandari2020zeromemoryoptimizationstraining} ZeRO Stage 3 as well as CPU offload, gradient clipping, and mixed precision; our DeepSpeed configuration is shown in Figure~\ref{fig:deepspeed}.

Parameters for SFT are shown in Table~\ref{tab:sft_config}. All SFT experiments use end-to-end fine-tuning instead of using parameter efficient methods such as LoRA~\citep{hu2021loralowrankadaptationlarge}. For model-generated reasoning traces, we use Qwen2.5-7B-Instruct as the teacher to keep the distribution of generations in the Qwen family. We use a temperature of 0.7 and filter out reasoning traces with 
the wrong answer. The resulting training set has 8,955 examples.

Versions of python and key libraries are shown in Table~\ref{tab:package-versions}. The dev version of trl was cloned directly from trl's GitHub repository on April 10, 2025.

\begin{table}[t]
\centering
\begin{tabular}{lc}
\toprule
Parameter & Value \\
\midrule
Learning rate                  & $2\times10^{-5}$ \\
Epochs              & 3               \\
Batch size per device & 4               \\
Gradient accumulation steps   & 2               \\
\bottomrule
\end{tabular}
\caption{Experimental configuration and hyperparameters for SFT.}
\label{tab:sft_config}
\end{table}

\begin{figure}
\begin{lstlisting}
compute_environment: LOCAL_MACHINE
debug: false
deepspeed_config:
  gradient_clipping: 1.0
  offload_optimizer_device: cpu
  offload_param_device: cpu
  zero3_init_flag: false
  zero3_save_16bit_model: false
  zero_stage: 3
distributed_type: DEEPSPEED
downcast_bf16: 'no'
enable_cpu_affinity: false
machine_rank: 0
main_training_function: main
mixed_precision: bf16
num_machines: 1
num_processes: 4
rdzv_backend: static
same_network: true
tpu_env: []
tpu_use_cluster: false
tpu_use_sudo: false
use_cpu: false
\end{lstlisting}
\caption{DeepSpeed configuration.}
    \label{fig:deepspeed}
\end{figure}

\paragraph{Coding.}
All experiments with MBPP+ on coding using on-policy RL for Qwen2.5-0.5B were conducted on AWS EC2 g6.12xlarge instances with 48 vCPUs, 192 GiB memory, and 4 NVIDIA L4 Tensor Core GPUs with 96 GiB total GPU memory, with 2 GPUs dedicated to training and 2 to sampling. Experiments with Qwen2.5-1.5B and Qwen2.5-3B were conducted on AWS EC2 g6e.12xlarge instances with 48 vCPUs, 384 GiB memory, and 4 NVIDIA L40S Tensor Core GPUs with 192 GB total GPU memory, with 2 GPUs dedicated to training and 2 to sampling. All SFT experiments on MBPP+ were conducted using 2 NVIDIA A6000 GPUs. The hyperparameters for coding are the same as for math.

\begin{table}[t]
\centering
\begin{tabular}{lc}
\toprule
Package & Version \\
\midrule
python & 3.11.11 \\
trl      & 0.17.0.dev0 \\
vllm     & 0.8.1        \\
pytorch  & 2.6.0        \\
\bottomrule
\end{tabular}
\caption{Package versions.}
\label{tab:package-versions}
\end{table}

\section{Additional Experiment Results}
\label{sec:additional_experiment}

\begin{figure}[t]
\centering
\begin{subfigure}[b]{0.4\textwidth}
\centering
\includegraphics[width=\textwidth]{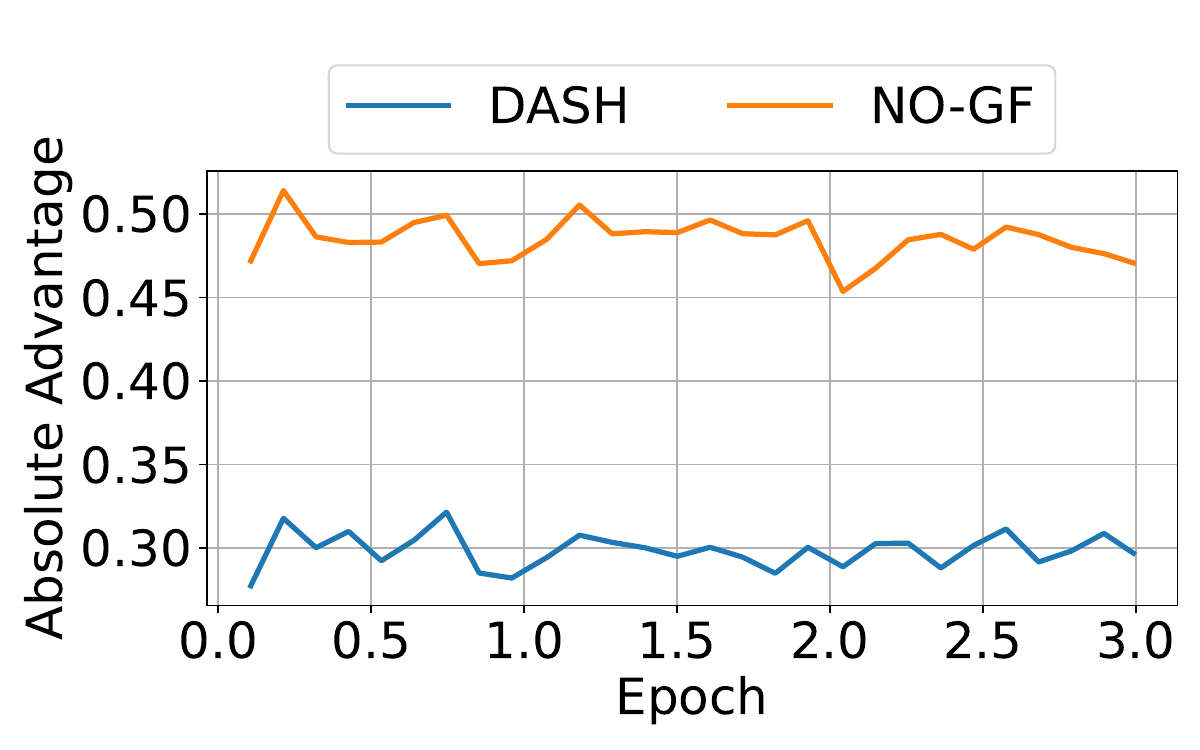}
\caption{Average absolute advantage values.}
\label{fig:mean_advantage_append}
\end{subfigure}
\
\begin{subfigure}[b]{0.4\textwidth}
\centering
\includegraphics[width=\textwidth]{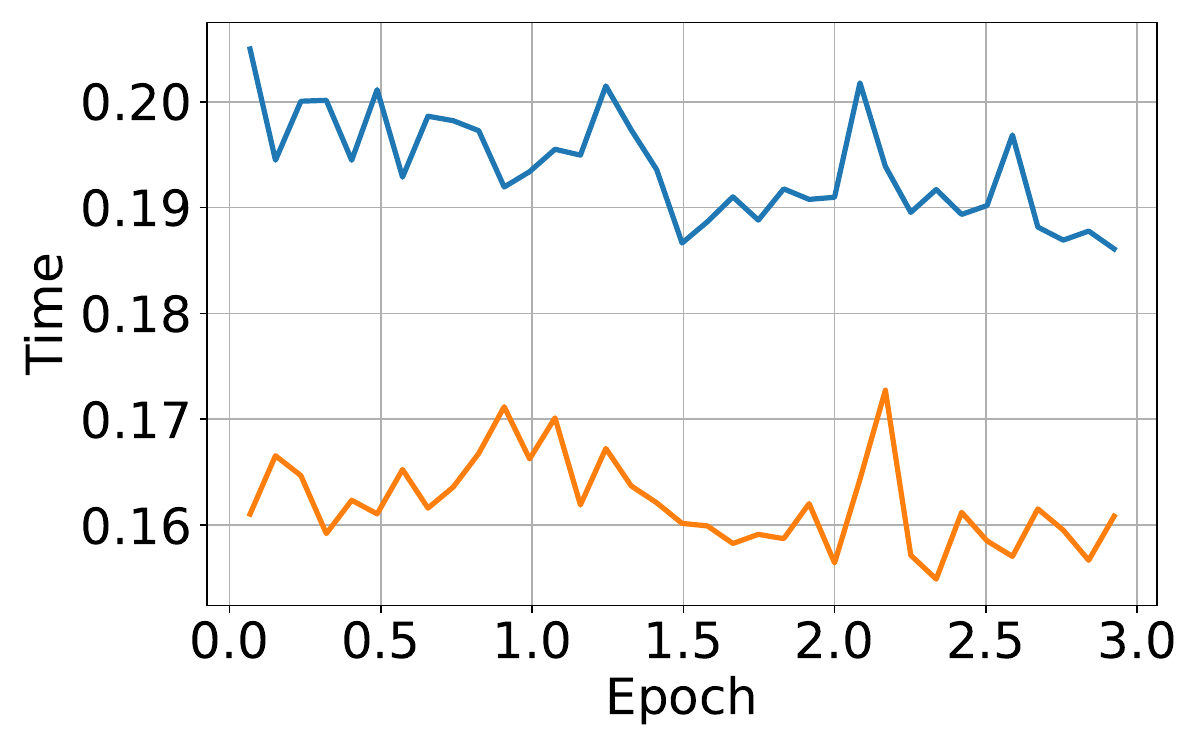}
\caption{Loss computation time.}
\label{fig:loss_computation_time_append}
\end{subfigure}
\
\begin{subfigure}[b]{0.4\textwidth}
\centering
\includegraphics[width=\textwidth]{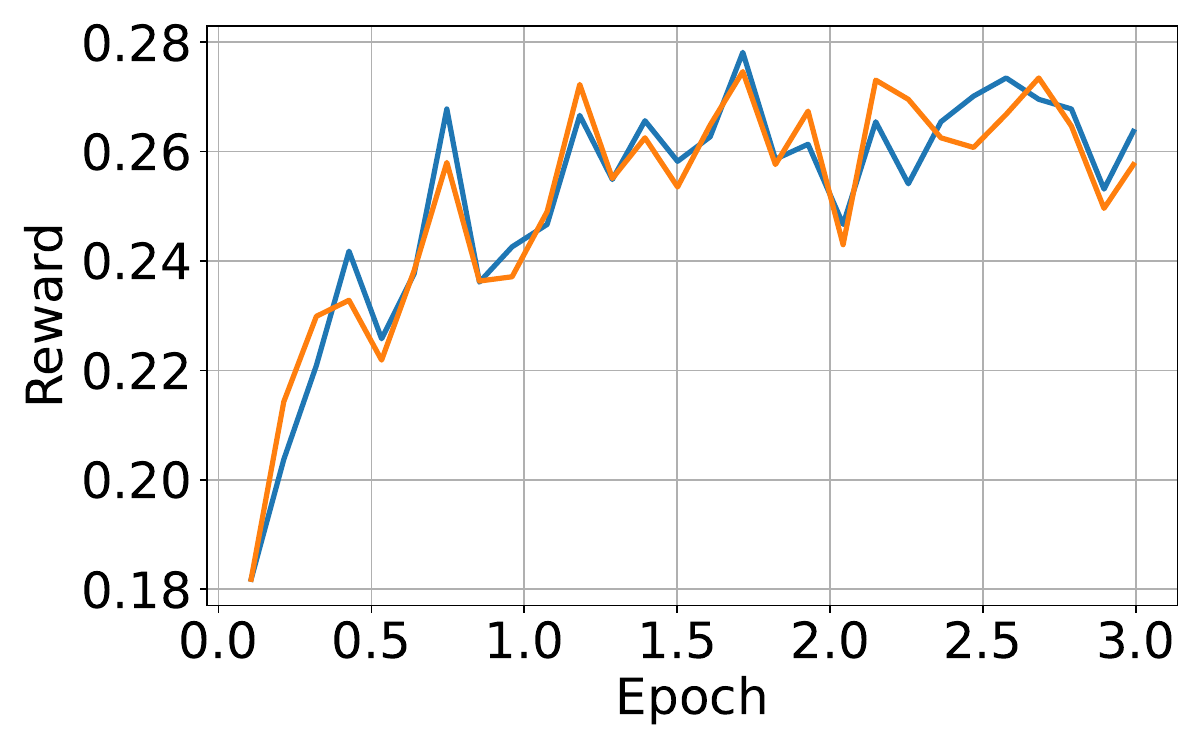}
\caption{Training reward curves.}
\label{fig:reward_gf_append}
\end{subfigure}
\caption{Comparison of No-GF and DASH with a per-device batch size of 4 for Qwen2.5-0.5B on math.}
\label{fig:three graphs_append}
\end{figure}

We compare gradient filtering with larger batch sizes, finding that gradient filtering is more effective when the per device batch size is 4 (instead of 2). This experiment is only possible for the  for Qwen2.5-0.5B on the math dataset using our compute. Results are shown in Table~\ref{tab:acceleration_app} and training curves are shown in Figure~\ref{fig:three graphs_append}. The time reduction achieved is larger than before (10\% instead of 4\%). These results suggest that gradient filtering may become more effective with larger batch sizes.

\begin{figure}[t]
\centering
\includegraphics[width=\linewidth]{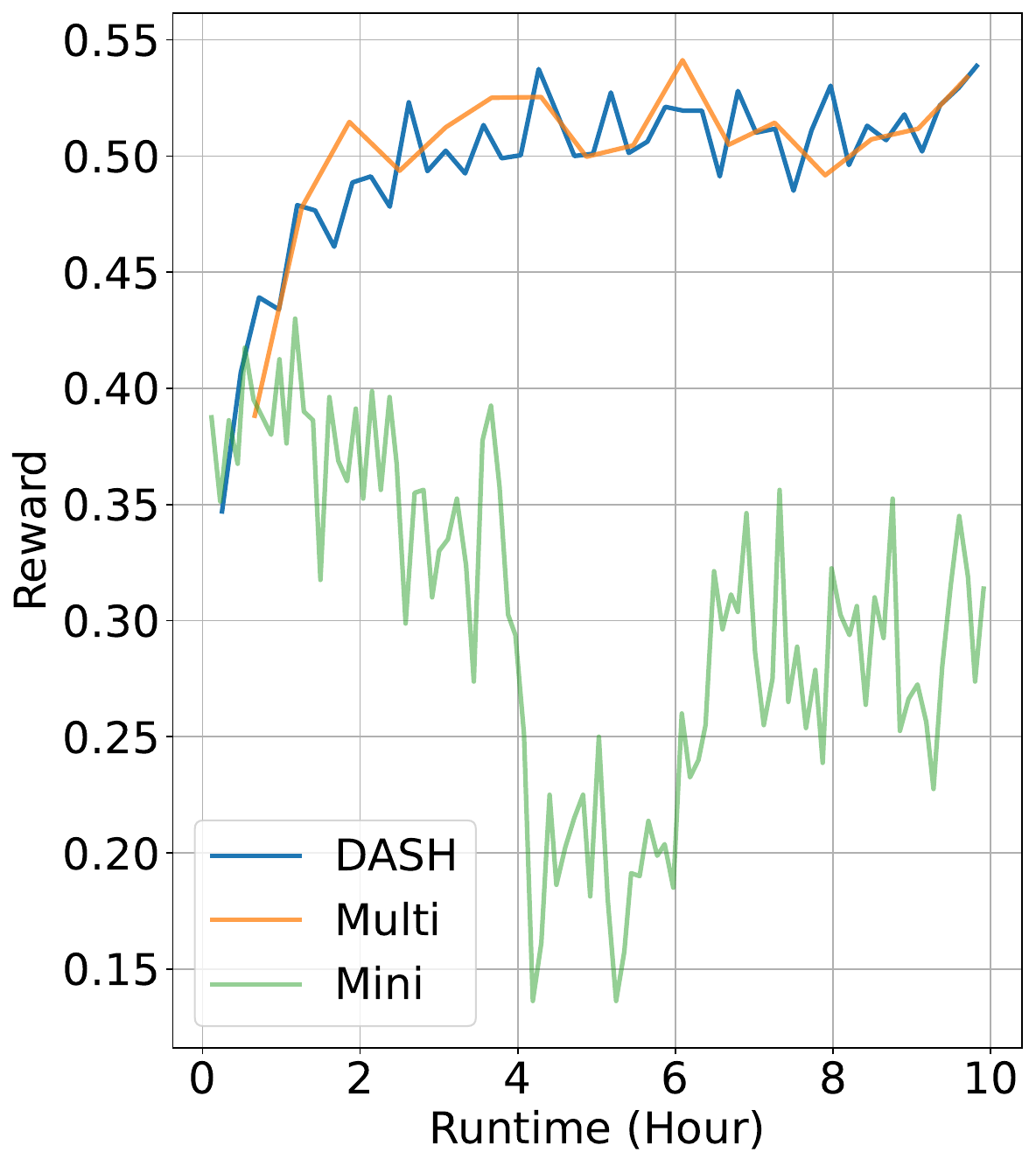}
\caption{Training reward curves for PG vs. PPO on Qwen2.5-1.5B for math.}
\label{fig:reward_curves_1.5B}
\end{figure}

We also show the comparison of Mini, Multi, and DASH on the 1.5B model (Figure~\ref{fig:reward_curves_1.5B}).  Conclusions are similar to Section~\ref{sec:gradient_update}. In this case we stabilize Multi by increasing the gradient accumulation step to 128, but the high instability of Mini leads to decrease in training rewards as well as accuracies as shown in Table~\ref{tab:grpo_comp}.

\begin{table}[t]
\centering
\begin{tabular}{cccc}\toprule
Method & Time (h) & MATH (\%) & GSM8K(\%)\\ \midrule
No-GF   &   5.1       &  31.8    & 31.3    \\ 
DASH   &   4.6       &  28.4    &  30.9   \\ \bottomrule
\end{tabular}
\caption{Comparing No-GF to DASH with per device batch of 4 instead of 2 for Qwen2.5-0.5B on math.}
\label{tab:acceleration_app}
\end{table}

\end{document}